\documentclass[runningheads]{llncs}
\usepackage{graphicx}

\usepackage{tikz}
\usepackage{comment}
\usepackage{amsmath,amssymb} 
\usepackage{color}

\usepackage{multirow}
\usepackage{graphicx}
\usepackage{amsmath}
\usepackage{amssymb}
\usepackage{booktabs}
\usepackage{float}
\usepackage{multirow}
\usepackage{subcaption}
\usepackage[super]{nth}
\usepackage{url}

\newcommand\ie{\emph{i.e.}} 
 
\newcommand\etal{\emph{et al.}}

\usepackage[pagebackref=true,breaklinks=true,letterpaper=true,colorlinks,bookmarks=false]{hyperref}

\begin{document}
\pagestyle{headings}
\mainmatter
\def\ECCVSubNumber{100}  

\title{An Improved RaftStereo Trained with A Mixed Dataset for the Robust Vision Challenge 2022} 

\titlerunning{2nd Place RVC 2022 - Stereo Matching}
%
\author{Hualie Jiang \and
Rui Xu \and
Wenjie Jiang}
%
\authorrunning{H. Jiang et al.}
%
\institute{
Insta360, Shenzhen 518000, China\\
\email{\{jianghualie, jerry1, jerett\}@insta360.com}}
\maketitle

\begin{abstract}
Stereo matching is a fundamental problem in computer vision. Despite recent progress by deep learning, improving the robustness is ineluctable when deploying stereo-matching models to real-world applications. Different from the common practices, i.e., developing an elaborate model to achieve robustness, we argue that collecting multiple available datasets for training is a cheaper way to increase generalization ability. Specifically, this report presents an improved RaftStereo~\cite{lipson2021raft} trained with a mixed dataset of \textbf{seven} public datasets for the robust vision challenge (denoted as iRaftStereo\_RVC). When evaluated on the training sets of Middlebury, KITTI-2015, and ETH3D, the model outperforms its counterparts trained with only one dataset, such as the popular Sceneflow. After fine-tuning the pre-trained model on the three datasets of the challenge, it ranks at \textbf{2nd place} on the stereo leaderboard\footnote{\url{http://www.robustvision.net/leaderboard.php?benchmark=stereo}}, demonstrating the benefits of mixed dataset pre-training. 
\end{abstract}

\section{Introduction}

Stereo matching is an everlasting research topic in computer vision, as it is a feasible way to recover the depth of environments and facilitates many downstream tasks, such as 3D reconstruction and robot navigation.
In recent years, research has shifted from traditional methods, such as SGM~\cite{hirschmuller2007stereo} and ELAS~\cite{geiger2010efficient} to deep learning based ones~\cite{sceneflow,iccv19_autodispnet,cheng2020hierarchical,xu2022attention,li2022practical,cheng2019learning,ganet,wang2021pvstereo,du2019amnet,zhang2020adaptive,lipson2021raft}. 
These deep-learning methods have considerably improved stereo matching. Most of these methods either regress the disparity from the 3D cost volume upon 2D features (AMNet~\cite{du2019amnet}, AcfNet~\cite{zhang2020adaptive},  OptStereo~\cite{wang2021pvstereo}, GANet-deep~\cite{ganet}, CSPN~\cite{cheng2019learning},  LEAStereo~\cite{cheng2020hierarchical}, and ACVNet~\cite{xu2022attention}), or directly from the 2D feature maps (DispNet~\cite{sceneflow}, AutoDispNet~\cite{iccv19_autodispnet}).

More recently, methods based on the strategy of excellent work on optical flow estimation, RAFT~\cite{teed2020raft}, iteratively update the disparity field using recurrent modules with the features sampled from the correlation volume, \ie, RaftStereo~\cite{lipson2021raft} and CREStereo~\cite{li2022practical}, establishing the state of the arts. 
However, although significant progress has been made on the model design, there is still a problem when deploying these models to diverse scenarios. This motivates the holding of the  Robust Vision Challenge, which encourages people to develop robust vision systems that work well in various scenes. 

To take part in the challenge, we first build the technical scheme upon RaftStereo~\cite{lipson2021raft}. 
In terms of robustness, Ranftl~\etal~\cite{ranftl2020towards} revealed that training with mixed datasets is powerful even for zero-shot cross-dataset generalization. 
We believe that it is also true for stereo matching. Therefore, we collect seven public available datasets, \ie, Sceneflow~\cite{sceneflow}, CreStereo~\cite{li2022practical}, Tartan~Air~\cite{tartanair2020iros}, Falling Things~\cite{tremblay2018falling}, Sintel~Stereo~\cite{butler2012naturalistic}, HR-VS~\cite{yang2019hsm} and InStereo2K~\cite{bao2020instereo2k}, and mix them for pre-training the RaftStereo~\cite{lipson2021raft}.  Experiments show that pre-training on the mixed dataset is beneficial compared with training on a single large dataset.  After fine-tuning the pre-trained model on the three training sets, it ranks at the \textbf{2nd place} on the stereo track of the challenge.

\section{Technical Scheme}
This section introduces the model and datasets we use in attending the challenge.

\subsection{RaftStereo}
RaftStereo~\cite{lipson2021raft} is a recently proposed stereo matching method, which follows the new paradigm of RAFT~\cite{teed2020raft} instead of regressing the disparity from the cost volume. Specifically, 
it iteratively updates the disparity field using recurrent modules with the features sampled from a 3D correlation pyramid. 
Different from RAFT which uses a single-level convolutional GRU, RaftStereo adopts multi-level convolutional GRUs for increasing the field of view faster.
We use the standard version of RaftStereo in the challenge, \ie, extracting the feature maps at 1/4 resolution and adopting 3-level GRUs. 
For further details, readers can refer to RaftStereo's paper and code\footnote{\url{https://github.com/princeton-vl/RAFT-Stereo}}.

\subsection{The Stereo Datasets}

This section presents the stereo datasets we use to construct the mixed dataset for pre-training a robust stereo-matching model.
Most of them are virtual datasets, as can be generated at scale. Table~\ref{tab:datasets} gives a summary of these datasets. 

\begin{table}[t]
  \centering
  \small
  \setlength{\tabcolsep}{5.pt}
  \begin{tabular}{l|c|c|c}
    \toprule
    Datasets & Types & \#Training Frames & Resolution \\
    \midrule
    \href{https://lmb.informatik.uni-freiburg.de/resources/datasets/SceneFlowDatasets.en.html#overview}{Sceneflow}~\cite{sceneflow} & Synthetic & 70908 & $540\times960$ \\
    \href{https://github.com/megvii-research/CREStereo#datasets}{CreStereo}~\cite{li2022practical} & Synthetic & 200000  & $1080\times1920$\\
    \href{https://github.com/castacks/tartanair_tools#download-the-training-data}{Tartan~Air}~\cite{tartanair2020iros} & Synthetic & 306637  & $480\times640$\\
    \href{https://research.nvidia.com/publication/2018-06_falling-things-synthetic-dataset-3d-object-detection-and-pose-estimation}{Falling Things}~\cite{tremblay2018falling} & Synthetic & 61500  & $540\times960$\\
    \href{http://sintel.is.tue.mpg.de/stereo}{Sintel~Stereo}~\cite{butler2012naturalistic} & Synthetic & 2128  & \ \ $436\times1024$\\
    \href{https://github.com/gengshan-y/high-res-stereo#data}{HR-VS}~\cite{yang2019hsm} & Synthetic & 780  & $2056\times2464$\\
    \href{https://github.com/YuhuaXu/StereoDataset#download}{InStereo2K}~\cite{bao2020instereo2k} & Realistic & 2010  & \ \ $860\times1080$\\
    \bottomrule
  \end{tabular}
  \vspace{10pt}
  \caption{Summary of the Stereo Datasets}
  \label{tab:datasets}
\end{table}

\noindent\textbf{Sceneflow:} Sceneflow~\cite{sceneflow} is a widely used dataset for stereo matching and optical flow estimation. It contains three subsets, i.e., FlyThings3D, Driving, and Monkaa. FlyThings3D is the biggest one, taking up over half of the dataset. The dataset is rendered with Blender~\cite{blender}, with a pixel resolution of 960x540. There are two types of passes for the rendered images, \ie, the clean pass and the final pass. Both are used in pre-training. 

\noindent\textbf{CreStereo:} CreStereo~\cite{li2022practical} is another synthetic dataset recently proposed by~Li~\etal, specifically for practical stereo matching. It is also rendered with Blender, but the object shapes, lighting and texture, and disparity distribution are different. The size of the dataset is larger than that of Sceneflow.

\noindent\textbf{Tartan~Air:} Tartan~Air~\cite{tartanair2020iros} is a photo-realistic dataset simulated with AirSim~\cite{airsim2017fsr} for visual SLAM of drones. It has a stereo track and provides depth maps, so we can use it in pre-training stereo-matching models. The dataset is even bigger than the former two, being the largest one.

\noindent\textbf{Falling Things:} 
Falling Things~\cite{tremblay2018falling} is a synthetic dataset with high graphical quality, for promoting object detection and 3D pose estimation methods in the context of robotics. As both stereo images and depth maps are available, we can use the datasets in pre-training a stereo disparity estimation network.

\noindent\textbf{Sintel-Stereo:} The Sintel dataset~\cite{butler2012naturalistic} is a popular synthetic dataset for  optical flow estimation. As the modalities, such as stereo images, disparity, and depth maps, are also provided, the dataset can be adopted in monocular and binocular depth estimation. Like Sceneflow, the dataset also provides two types of passes, \ie, the clean pass and the final pass. Although the dataset is relatively small, it presents dynamic humans and animals. We include it for potentially better generalization.

\noindent\textbf{HR-VS:} HR-VS~\cite{yang2019hsm} is a relatively small dataset rendered by the Carla simulator~\cite{Dosovitskiy17}, having only 780 stereo pairs. It is designed for high-resolution stereo matching, with a resolution of 2056$\times$2464. The scenarios are about driving, so we include it for better generalizing to driving scenes.

\noindent\textbf{InStereo2K:} InStereo2K~\cite{bao2020instereo2k} is a real-scene indoor stereo dataset, containing about 2k pairs of images with high-accuracy disparity maps. It is proposed to improve the performance of deep stereo-matching networks in practical applications. Therefore, we include it with the synthetic one in the mixed dataset.

To reduce the imbalance of these datasets, we copy Sceneflow, Falling Things, Sintel-Stereo, HR-VS, and InStereo2K 3, 3, 10, 25, and 10 times before adding them into the mixed dataset. We use the mixed dataset for pre-training. 
After that, we finetune the pre-trained model on the three stereo datasets of the challenge, \ie, KITTI-2015~\cite{menze2015object}, Middlebury~\cite{middlebury}, and ETH3D~\cite{eth3d}.
The training sets of these datasets have 200, 15, and 153 pairs (We also include the Middlebury 2014~\cite{scharstein2014high} for training). To improve the balance, we repeat ETH3D 10 times in finetuning.

\subsection{Training Schedule}

We perform our experiments upon the open-source code of RaftStereo~\cite{lipson2021raft}, which is implemented in Pytorch~\cite{pytorch}. Most of the training settings are the same as the implementation of RaftStereo. The differences lie in the input size and batch size. We use a batch size of 4 instead of 8, as we use two RTX 2080Ti GPUs for training, which have less memory than the RTX 6000 GPUs. Meanwhile, we also reduce the input size from $360\time720$ to $320\time704$ to save the memory. 
Therefore, doing spatial data augmentation, we random crop a region of $320\time704$ in all experiments. The remaining data augmentation is the same to RaftStereo~\cite{lipson2021raft}.

To participate in the challenge, we train the model in two phases. 
In the first phase, we train the model with the mixed dataset. We also have experimented with Sceneflow and CreStereo datasets for comparison. 
In the fine-tuning stage, we train the combination of three training sets as suggested by the challenge. We use different minimum learning rates in two phases, \ie, $1e^{-4}$ for the pre-training, and  $1e^{-5}$ for fine-tuning. In pre-training, we conventionally train the model with 200k steps. In fine-tuning, there are only 30k iterations.

\begin{figure*}[!ht]
\centering
\resizebox{\textwidth}{!}{
\newcommand{\turnheightnew}{0.20\columnwidth}
\newcommand{\turnwidthnew}{0.45\columnwidth}

\centering

\renewcommand{\arraystretch}{0.5}
\begin{tabular}{@{}c@{\hskip 0.5mm}c@{\hskip 0.5mm}c@{\hskip 0.5mm}c@{\hskip 0.5mm}c@{}}

{\rotatebox{90}{\hspace{3mm} Left Image}} &
\includegraphics[height=\turnheightnew, width=\turnwidthnew]{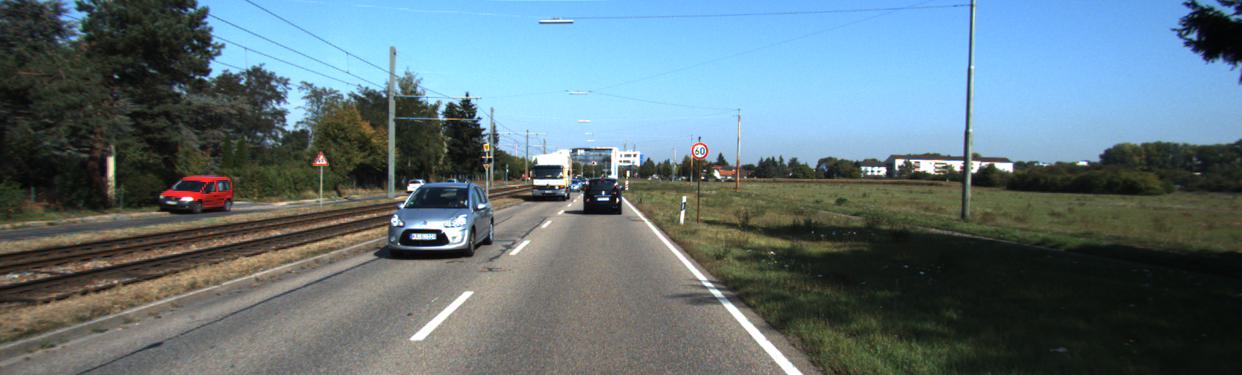} &
\includegraphics[height=\turnheightnew, width=\turnwidthnew]{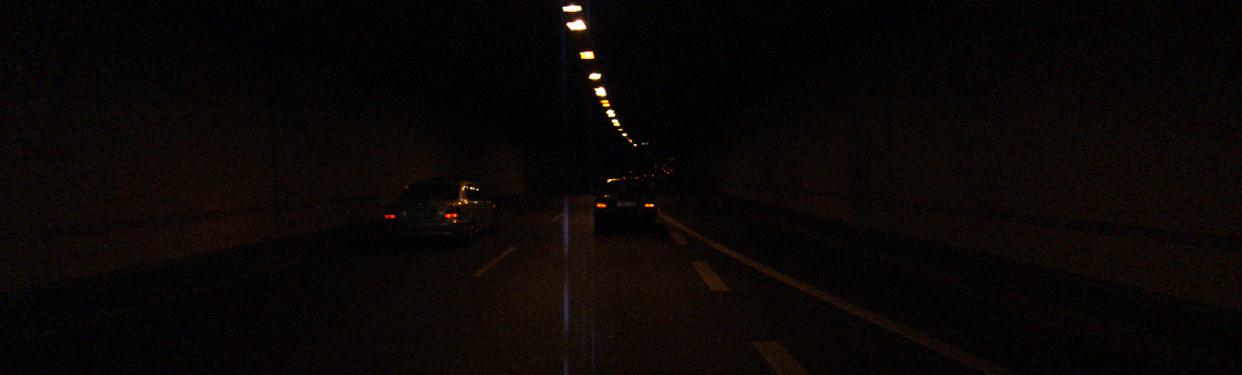} &
\includegraphics[height=\turnheightnew]{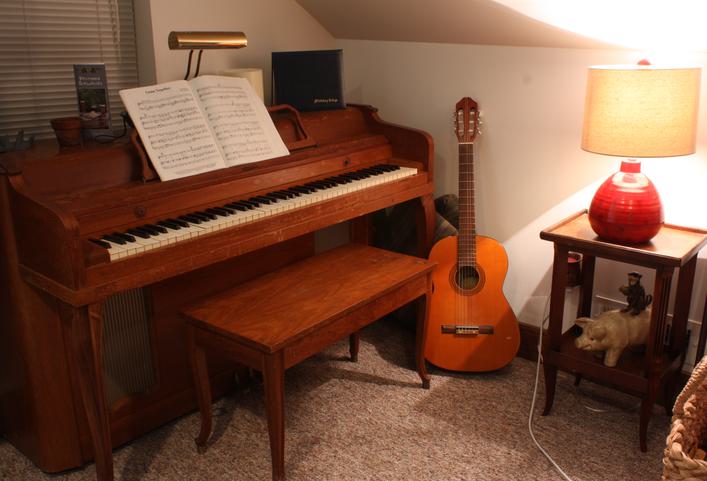} &
\includegraphics[height=\turnheightnew]{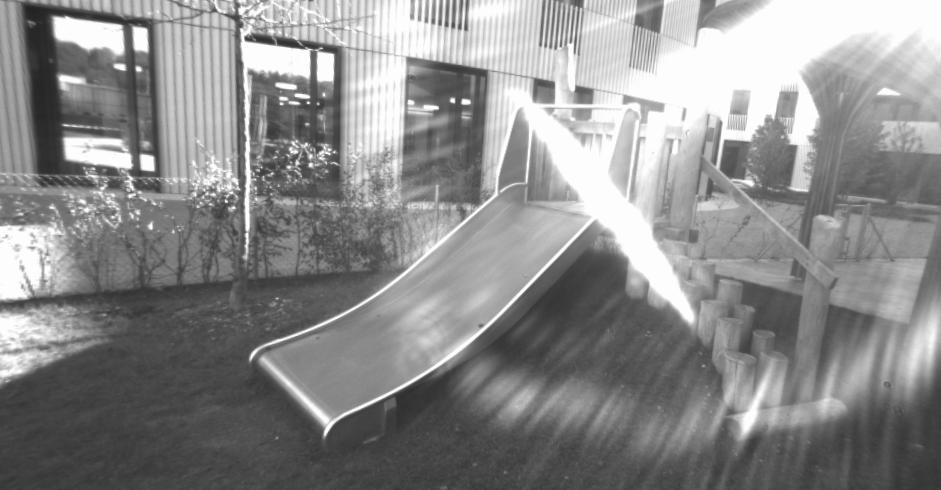} \\

{\rotatebox{90}{\hspace{1mm} Ground Truth}} &
\includegraphics[height=\turnheightnew, width=\turnwidthnew]{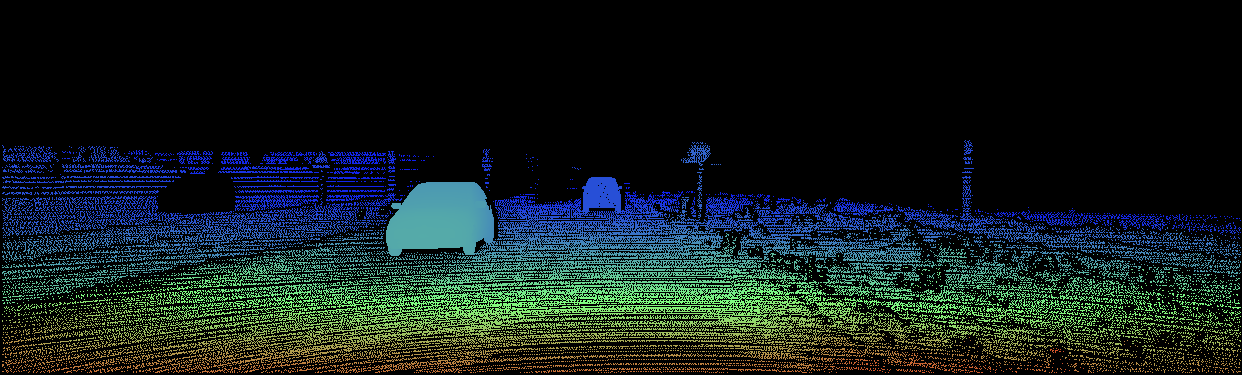} &
\includegraphics[height=\turnheightnew, width=\turnwidthnew]{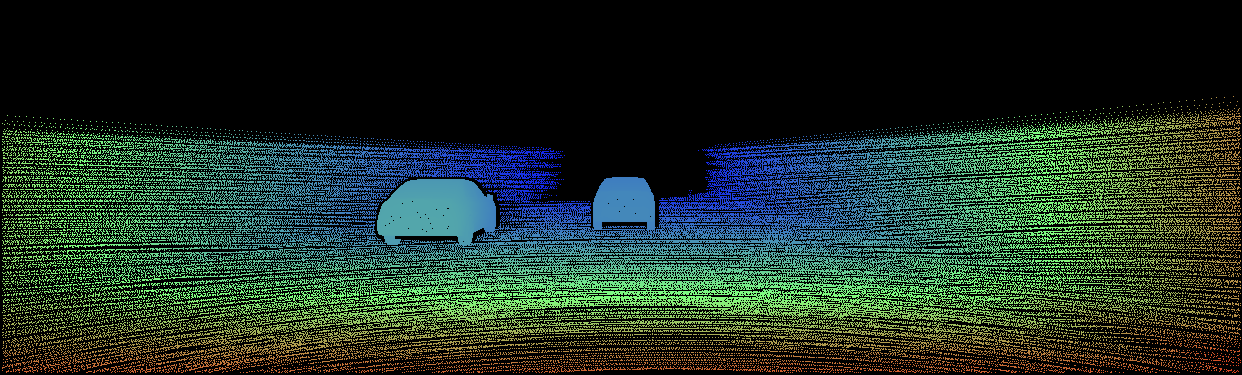} &
\includegraphics[height=\turnheightnew]{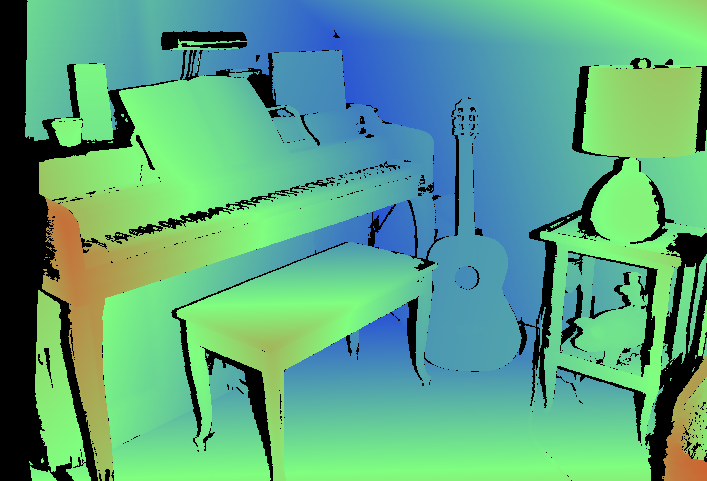} &
\includegraphics[height=\turnheightnew]{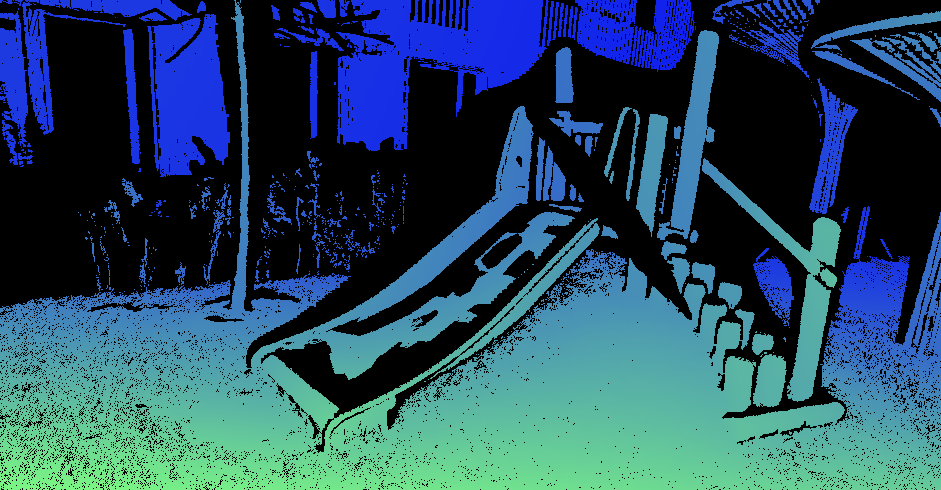} \\

{\rotatebox{90}{\hspace{0mm} Mixed Datasets}} &
\includegraphics[height=\turnheightnew, width=\turnwidthnew]{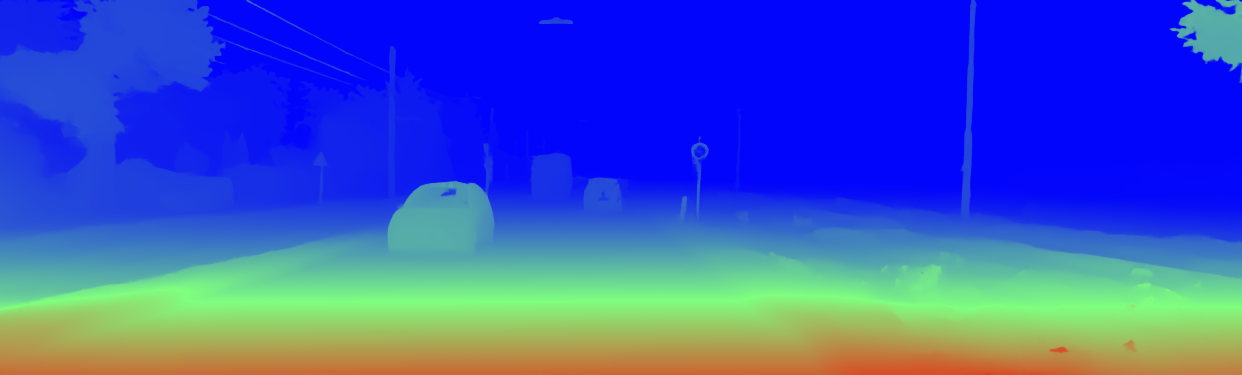} &
\includegraphics[height=\turnheightnew, width=\turnwidthnew]{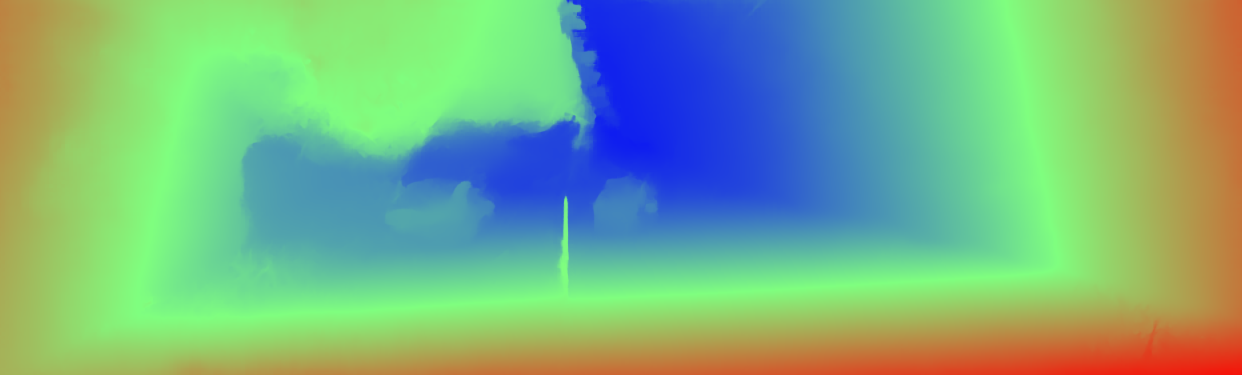} &
\includegraphics[height=\turnheightnew]{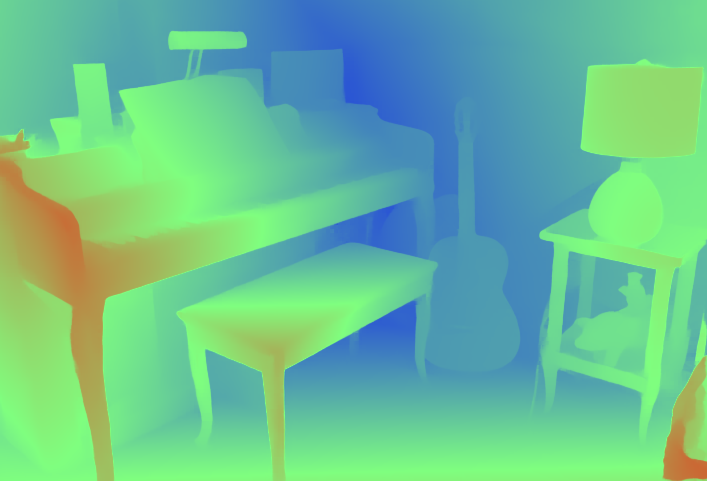} &
\includegraphics[height=\turnheightnew]{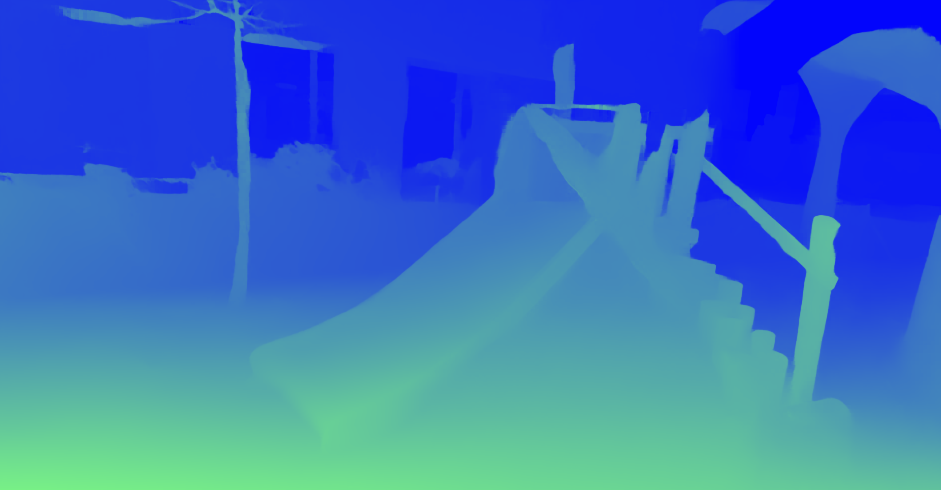} \\

{\rotatebox{90}{\hspace{3mm} SceneFlow}} &
\includegraphics[height=\turnheightnew, width=\turnwidthnew]{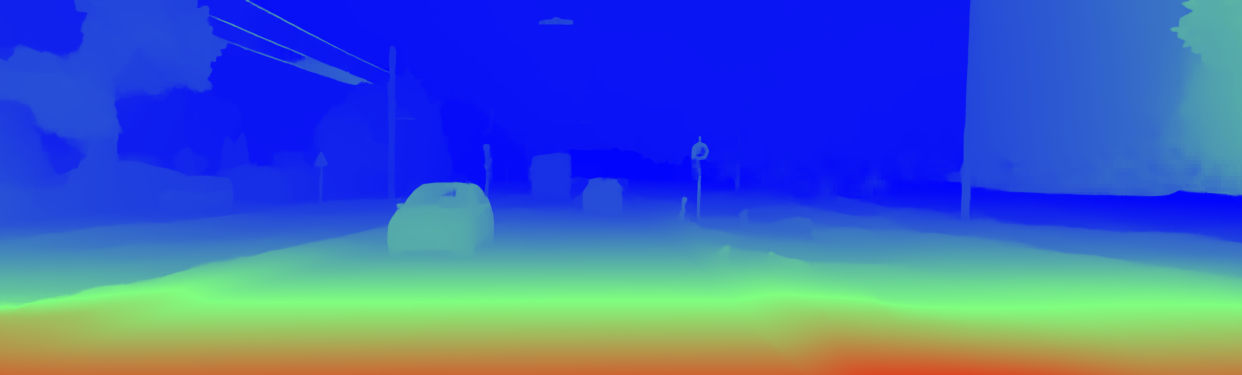} &
\includegraphics[height=\turnheightnew, width=\turnwidthnew]{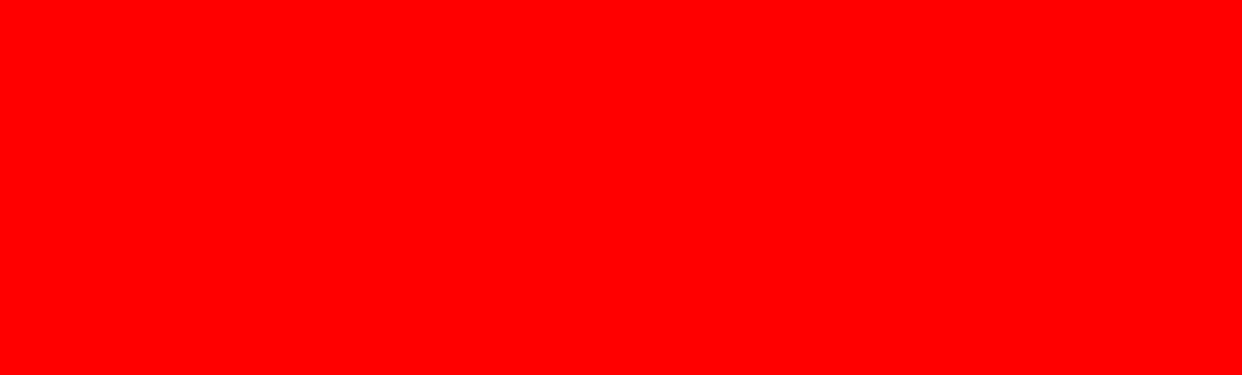} &
\includegraphics[height=\turnheightnew]{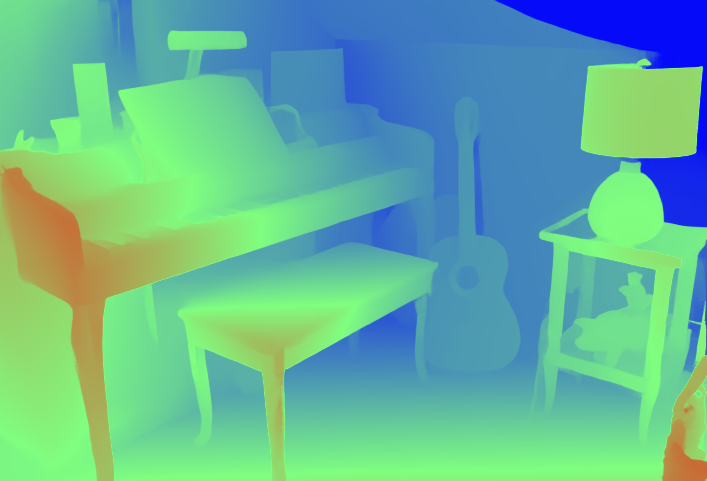} &
\includegraphics[height=\turnheightnew]{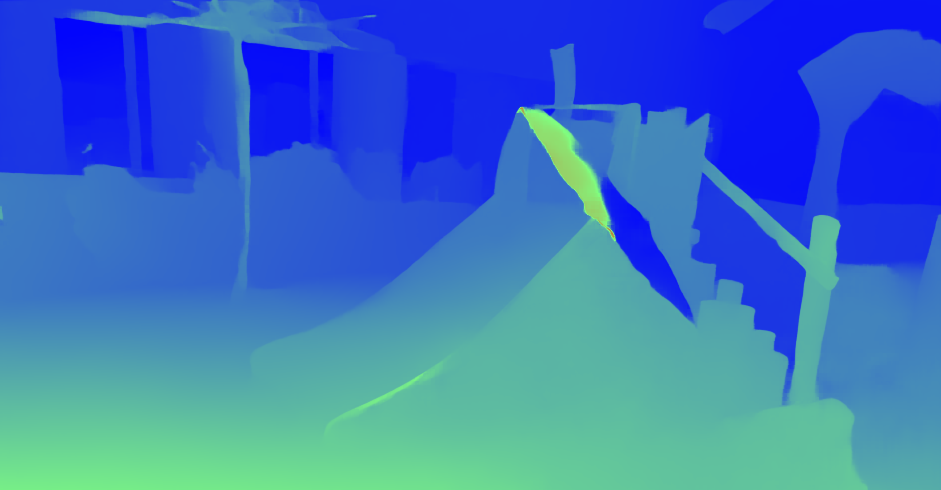} \\

{\rotatebox{90}{\hspace{4mm} CreStereo}} &
\includegraphics[height=\turnheightnew, width=\turnwidthnew]{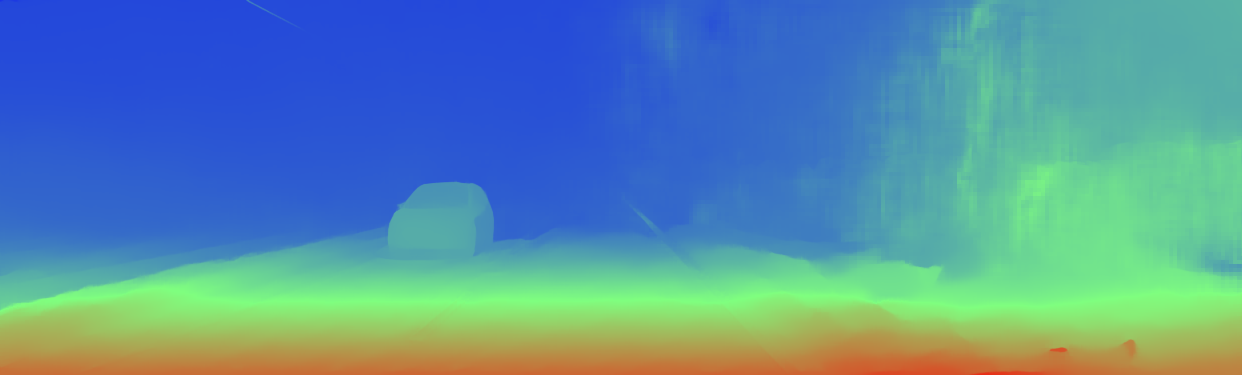} &
\includegraphics[height=\turnheightnew, width=\turnwidthnew]{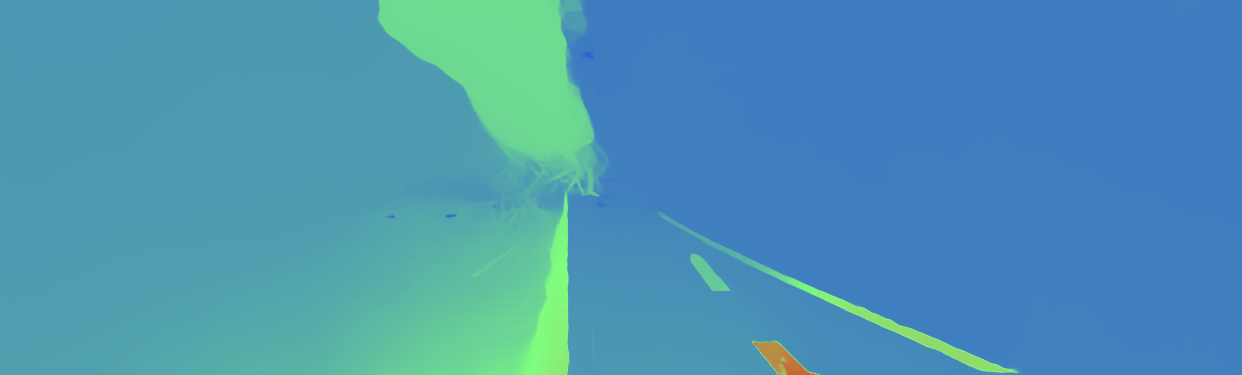} &
\includegraphics[height=\turnheightnew]{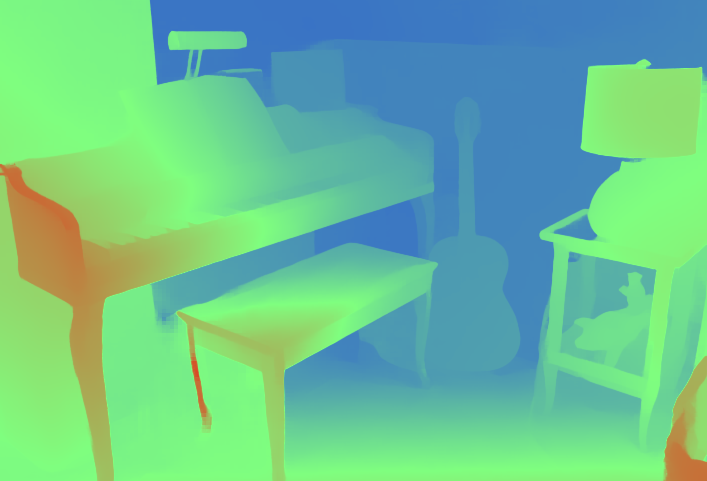} &
\includegraphics[height=\turnheightnew]{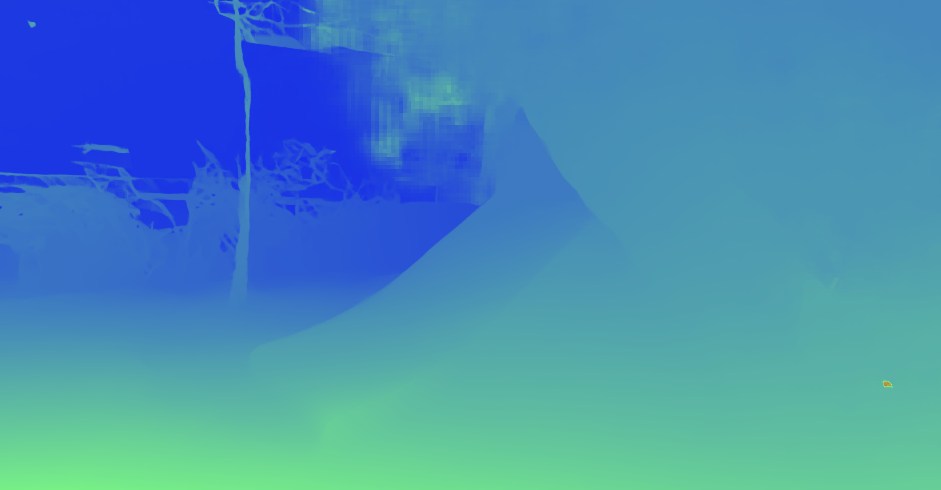} \\

 &
KITTI-2015 & 
KITTI-2015 &
Middlebury &
ETH3D\\

\end{tabular}

 }
\caption{\textbf{Qualitative comparison of the pre-training models.}}
\label{fig:comparison}
\end{figure*}

\section{Results}

The section presents the zero-shot generalization results of the pre-trained models, and the benchmark results on KITTI-2015 of the fine-tuned model. 

\subsection{Pre-training Results}

\begin{table}[t]
  \centering
  \small
  \setlength{\tabcolsep}{3.pt}
  \begin{tabular}{lcccccccc}
    \toprule
    \multirow{3}{*}[-2pt]{\shortstack{Pre-training\\Datasets}} &  \multicolumn{2}{c}{\multirow{2}{*}{KITTI-2015}} &  \multicolumn{4}{c}{Middlebury} & \multicolumn{2}{c}{\multirow{2}{*}{ETH3D}}\\ 
    \addlinespace[-12pt] \\
    & & & \multicolumn{2}{c}{Half}  & \multicolumn{2}{c}{Quarter} & & \\
    \cmidrule(lr){2-3} \cmidrule(lr){4-5} \cmidrule(lr){6-7} \cmidrule(lr){8-9}
    \addlinespace[-12pt] \\ 
    & bad 3.0 & avgerr & bad 2.0 & avgerr & bad 2.0 & avgerr & bad 1.0 & avgerr\\
    \midrule
   Mixed Dataset & \textbf{5.49} & \textbf{1.10} & \textbf{10.04} & \underline{1.64} & \textbf{8.44} & \textbf{0.98} & \textbf{2.59} & \textbf{0.25} \\
   SceneFlow & \underline{6.30} & 1.77 & \underline{11.89} & \textbf{1.59} & \underline{8.59} & \underline{1.01} & \underline{3.42} & 0.30\\
   CreStereo & 7.60 & \underline{1.30} & 18.54 & 2.56 & 15.45 & 1.41 & 4.20 & \underline{0.29} \\
    \bottomrule
  \end{tabular}
  \vspace{10pt}
  \caption{The evaluation results of models pre-trained on 3 different datasets. The best one is in \textbf{bold}, and the second best one is \underline{underline}.}
  \label{tab:pre-traing}
\end{table}

Table~\ref{tab:pre-traing} show the zero-shot generalization results on the training sets of KITTI-2015, Middlebury, and ETH3D with models pre-trained on different datasets. Due to the memory limitation of the GPU, we cannot evaluate Middlebury under full resolution. Generally, the model pre-training the adopted mixed dataset performs quantitatively better than its two counterparts. 
We also find that the model pre-trained the mixed dataset is more robust to common factors, such as non-textures and overexposure, as illustrated in Fugure~\ref{fig:comparison}.

\subsection{Benchmark Results}

After fine-tuning the pre-training model on training sets of the challenge, we submitted the predictions of the test sets of KITTI-2015, Middlebury, and ETH3D, to their online evaluation sites, separately. Our model, iRaftStereo\_RVC, finally ranks at 2nd place overall. Our model ranks 2nd on Middlebury and ETH3D. 

iRaftStereo\_RVC only ranks 4th on the KITTI-2015 in the challenge, and it appears to lag. 
However, after examining the detailed detailed quantitave results, we find another practical merit of the model compared with other models. iRaftStereo\_RVC relatively presents lower errors on the foreground regions. 
As pointed out by Jinag~\etal~\cite{jiang2021unsupervised}, the accuracy of foreground objects is more critical for practical application, but the depth estimation usually suffers from the foreground region more than the background region.

\begin{table}[t]
  \centering
  \small
  \setlength{\tabcolsep}{5.pt}
  \begin{tabular}{lcccc}
    \toprule
    Method & all & backgr. & foregr. & foregr./backgr.\\
    \midrule
    LEAStereo~\cite{cheng2020hierarchical} & \textbf{1.65}  & \underline{1.40} & 2.91 &2.08 \\
    ACVNet~\cite{xu2022attention} & \textbf{1.65} & \textbf{1.37}   & 3.07  & 2.24  \\
    CREStereo~\cite{li2022practical} &1.69 &1.45 &2.86  &1.97 \\
    CSPN~\cite{cheng2019learning} & 1.74 & 1.51 & \underline{2.88} & \underline{1.91} \\
    GANet-deep~\cite{ganet} & 1.81 & 1.48 & 3.46 & 2.34 \\
    OptStereo~\cite{wang2021pvstereo} & 1.82 & 1.50 & 3.43 & 2.29 \\
    AMNet~\cite{du2019amnet} & 1.84 & 1.53  & 3.43 & 2.24 \\
    AcfNet~\cite{zhang2020adaptive} & 1.89 & 1.51  & 3.80 & 2.52\\
    RaftStereo~\cite{lipson2021raft} & 1.96 & 1.75  & 2.89  & \textbf{1.65} \\
    \midrule
    raft+\_RVC   &\textbf{1.83}  &\underline{1.60} &\textbf{2.98}  &1.86 \\
    CREStereo++\_RVC        &\underline{1.88}  &\textbf{1.55} &3.53  &2.28 \\     
    MaskLacGwcNet\_RVC      &1.99  &1.65 &3.68  &2.23 \\            
    iRaftStereo\_RVC(\textbf{Ours}) &2.07  &1.88 &\underline{3.03}  &\textbf{1.61} \\
    GANetREF\_RVC           &2.33  &1.88 &4.58  &2.44 \\          
    CroCo\_RVC              &2.33  &2.04 &3.75  &\underline{1.84} \\       
    GEStereo\_RVC           &2.71  &2.29 &4.79  &2.09 \\       
    SGM\_RVC                &6.38  &5.06 &13.00 &2.57 \\          
    ELAS\_RVC               &9.67  &7.38 &21.15 &2.87 \\       
    \bottomrule
  \end{tabular}
  \vspace{10pt}
  \caption{The KITTI-2015 benchmark results of some state-of-the-art methods and the participators of the challenge. The metirc is the percentage of erroneous (EPE $>$ 3 px) pixels, \ie, bad 3.0. The best one is in \textbf{bold}, and the second best one is \underline{underline}.}
  \label{tab:kitti}
\end{table}

Table~\ref{tab:kitti} lists the KITTI-2015 benchmark results of some state-of-the-art methods and the participators of the challenge. 
Among the participators, iRaftStereo\_RVC has the 2nd smallest bad 3.0 on the foreground region, just slightly bigger than that of the best one, raft+\_RVC (3.03 vs. 2.98). 
We also compute the ratio of the errors between the foreground and the background in Table~\ref{tab:kitti}. 
Most methods tend to overfit the background, as their foregr./backgr. ratios approximate to 2 or probably are higher. RaftStereo~\cite{lipson2021raft} produces the smallest ratio, 1.65, among the published methods, indicating it balances the foreground and background better. Our iRaftStereo\_RVC has an even lower ratio, 1.61, thus the mixed dataset pre-training does not sacrifice RaftStereo's advantage in balancing the foreground and background. In contrast, raft+\_RVC, probably a modified version of RaftStereo, increases the ratio to 1.86, although it achieves the best overall result in the challenge

\begin{figure*}[!ht]
\centering
\resizebox{\textwidth}{!}{
\newcommand{\turnheightnew}{0.20\columnwidth}

\centering

\renewcommand{\arraystretch}{0.5}
\begin{tabular}{@{\hskip 1mm}c@{\hskip 1mm}c@{\hskip 1mm}c@{}}

{\rotatebox{90}{\hspace{2.5mm} Left Image}} &
\includegraphics[height=\turnheightnew]{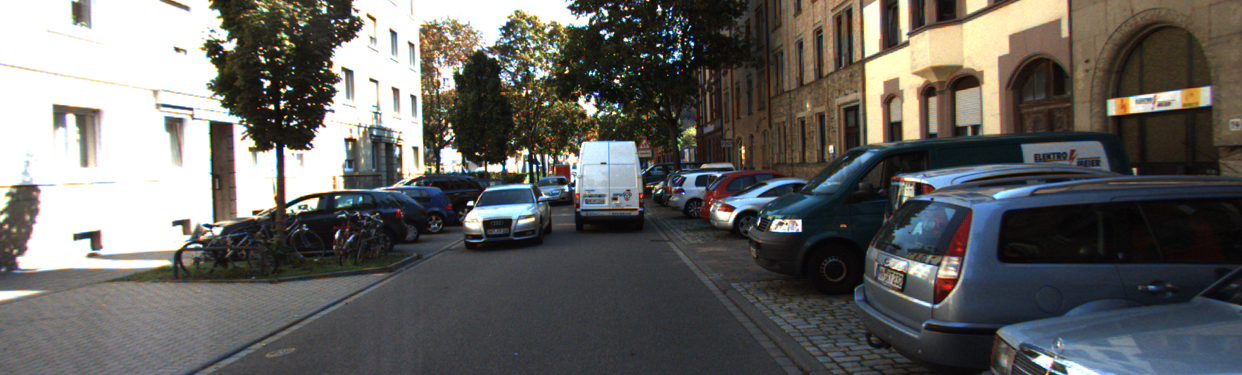} &
\includegraphics[height=\turnheightnew]{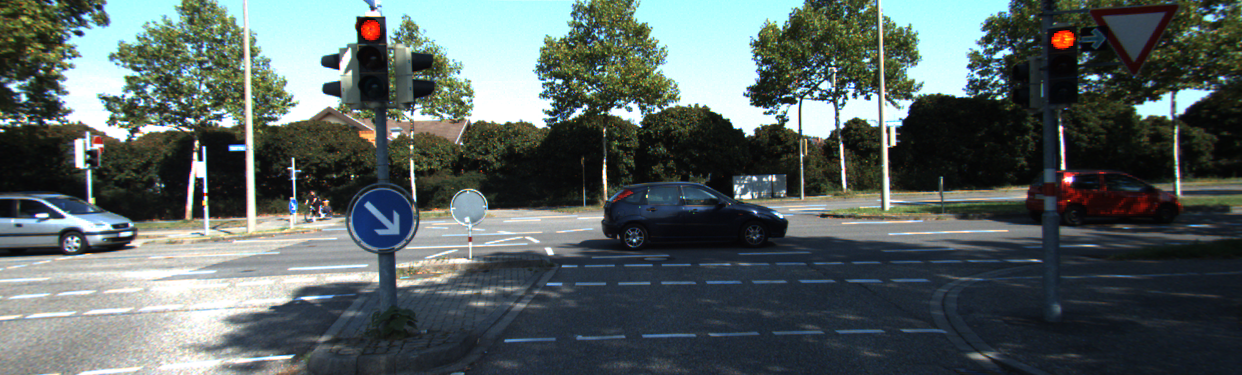} \\

{\rotatebox{90}{\hspace{2.5mm} raft+\_RVC}} &
\includegraphics[height=\turnheightnew]{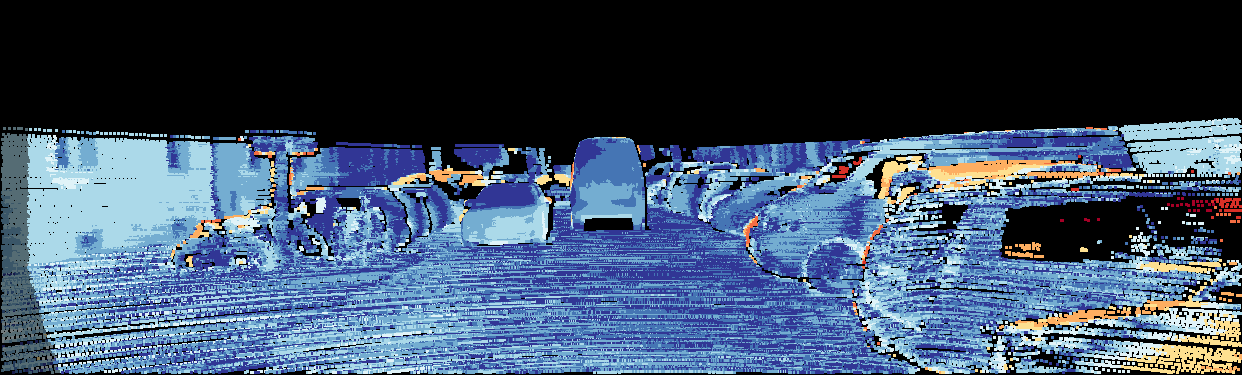} &
\includegraphics[height=\turnheightnew]{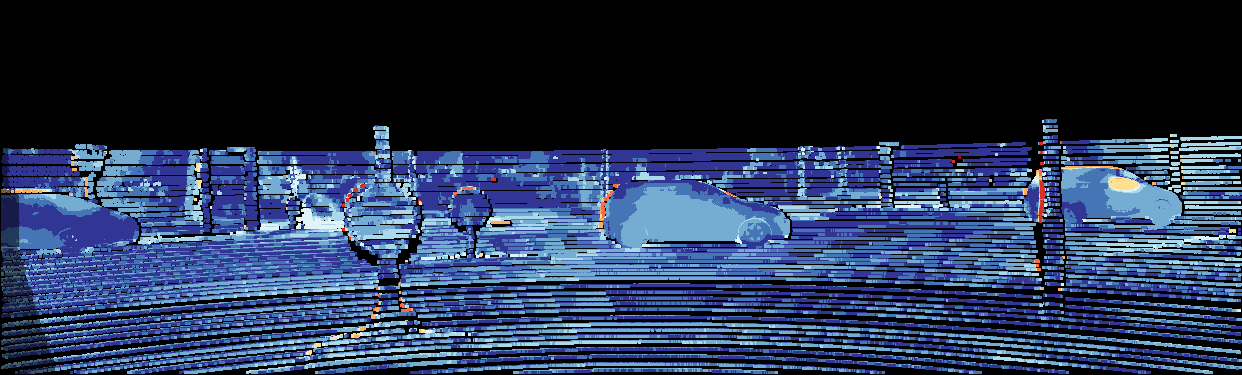} \\

{\rotatebox{90}{\hspace{-1mm} \scriptsize CREStereo++\_RVC}} &
\includegraphics[height=\turnheightnew]{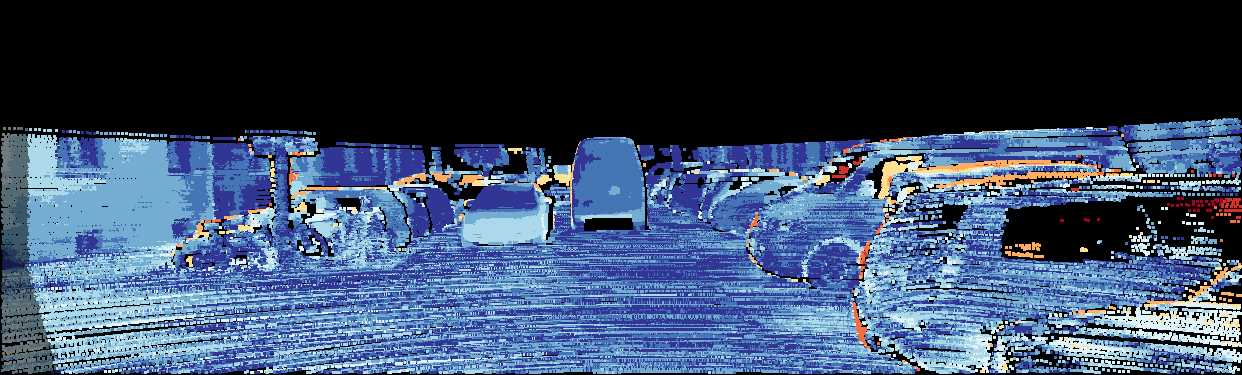} &
\includegraphics[height=\turnheightnew]{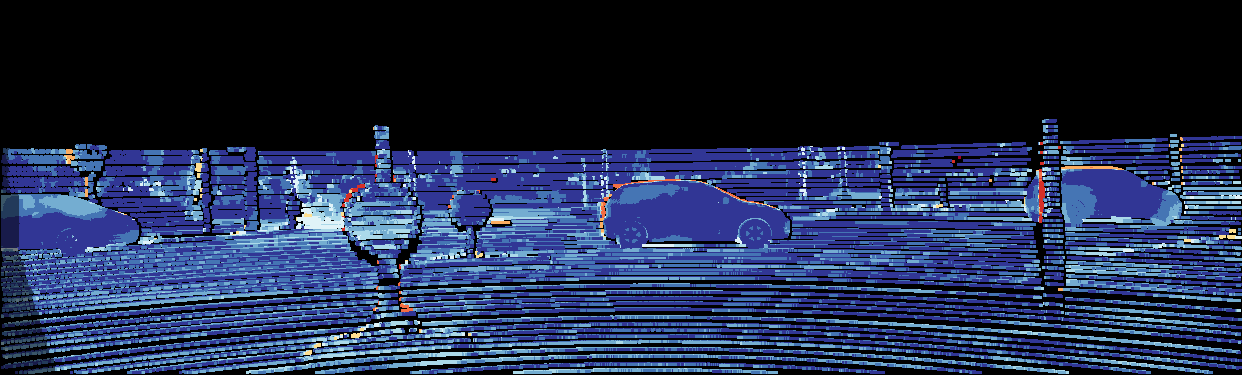} \\

{\rotatebox{90}{\hspace{0.1mm} \scriptsize iRfatStereo\_RVC}} &
\includegraphics[height=\turnheightnew]{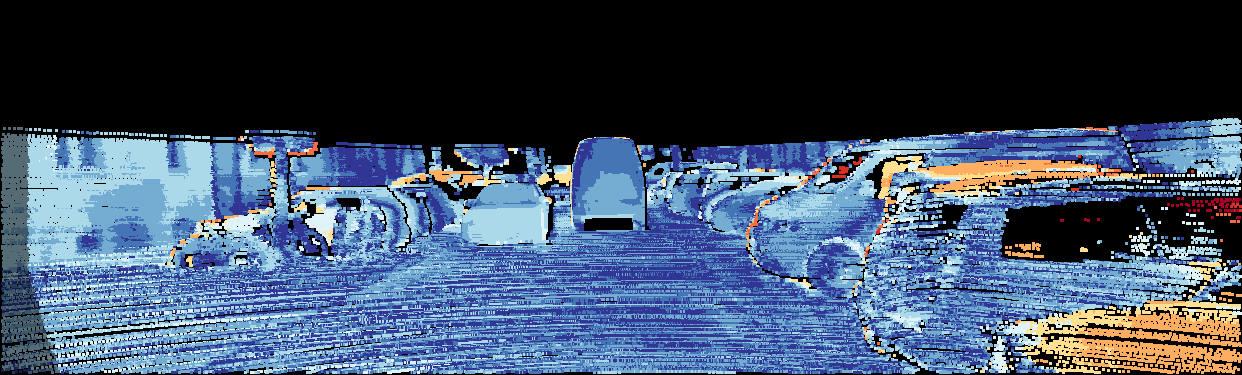} &
\includegraphics[height=\turnheightnew]{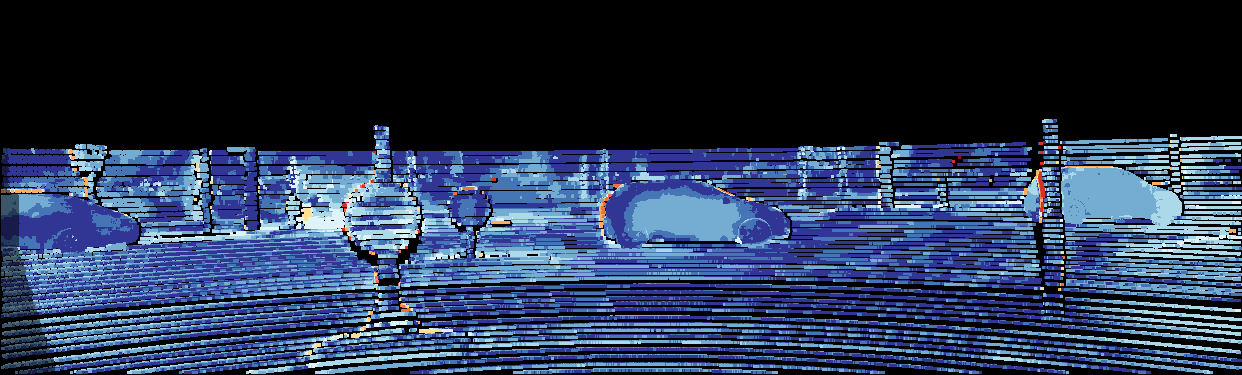} \\

\end{tabular}
 }
\caption{\textbf{The comparison of KITTI-2015 error maps of three participators of the challenge.}}
\label{fig:error_maps}
\end{figure*}

We also collect some error maps from KITTI evaluations sites\footnote{\url{https://www.cvlibs.net/datasets/kitti/eval_scene_flow.php?benchmark=stereo}} for a better comparison, which are shown in Figure~\ref{fig:error_maps}. One can see that, raft+\_RVC and iRaftStereo\_RVC present lower errors on the front vehicles than CREStereo++\_RVC. 

\newpage

\hspace{5pt}

\section{Conclusions}

We have presented our scheme for participating in the stereo track of the Robust Vision Challenge 2022. We adopted the state-of-the-art model, RaftStereo, and collect a mixed dataset of seven different datasets in pre-training. Experiments have shown that compared to conventionally pre-training on a single large dataset, pre-training on the mixed dataset makes the model robust to textureless and over-exposure regions, and produces better results in zero-shot generalization. After fine-tuning the pre-trained model on the training sets of the challenge, we rank in 2nd place. Although the ranking of KITTI-2015 is 4th, after inspecting the results deeper, we found our model is very advantageous in handling the foreground objects, which are more critical than the background in practical applications. To sum up, mixed dataset training helps to achieve robustness. 

{\paragraph{\textbf{Acknowledgments:}} We would like to thank the authors of RaftStereo, as our technical scheme depends on their excellent work.}

\clearpage
%
%
\bibliographystyle{splncs04}
\bibliography{egbib}
\end{document}